\pdfoutput=1

\documentclass[11pt]{article}

\usepackage[T2A,T1]{fontenc}
\usepackage[]{acl}
\usepackage{listings}
\usepackage{times}
\usepackage{latexsym}
\usepackage{algorithm}
\usepackage{algpseudocode}
\usepackage{booktabs}
\usepackage{array}
\usepackage{ipa}
\usepackage{stackengine}
\usepackage{wrapfig,lipsum,booktabs}
\newcolumntype{P}[1]{>{\centering\arraybackslash}p{#1}}
\usepackage{longtable}
\usepackage{scalefnt}
\usepackage{graphicx}

\usepackage{multicol}
\usepackage{multirow}
\usepackage[T1]{fontenc}

\usepackage[utf8]{inputenc}

\usepackage{CJKutf8}

\usepackage{microtype}
\usepackage{colortbl}
\usepackage{arydshln}
\usepackage{stfloats}
\usepackage{booktabs}

\definecolor{lgreen}{RGB}{73,174,137}
\definecolor{lred}{RGB}{182,49,54}
\definecolor{lorange}{RGB}{255, 128, 0}
\definecolor{lblue}{RGB}{0, 0, 255}

\newcommand{\enxx}{en$\rightarrow$xx }
\newcommand{\xxen}{xx$\rightarrow$en }

\newcommand{\chrf}{\textsc{ChrF}}
\newcommand{\gatitos}{\mbox{\textsc{Gatitos}}}
\newcommand{\smol}{\textsc{Smol}}
\newcommand{\smoldog}{\textsc{SmolDoc}}
\newcommand{\smoldogsent}{\textsc{SmolDocSplit}}
\newcommand{\smolseal}{\textsc{SmolSent}}
\newcommand{\smolboth}{\textsc{Both}}
\newcommand{\smolbothg}{\textsc{Both+G}}
\newcommand{\nsmolnonenglishlangs}{124} 
\newcommand{\nsmolnonenglishps}{125} 
\newcommand{\nsmolsealnonenglishlangs}{88} 
\newcommand{\nsmoldognonenglishlangs}{106} 
\newcommand{\nsmoldogdocs}{584}
\newcommand{\nllbseed}{\textsc{NLLB-Seed}}
\newcommand{\floresone}{\textsc{Flores-101}}
\newcommand{\florestwo}{\textsc{Flores-200}}
\newcommand{\bread}{\textsc{Bread}}
\newcommand{\ntrex}{\textsc{NTreX}}

\newcommand{\madlad}{\textsc{Madlad-400}}

\newcommand{\ugh}{$^{\textrm{\small \revglotstop}}$}
\newcommand{\oho}{$^\textrm{\small \openo}$}
\newcommand{\otho}{$^\textrm{\small \dh}$}
\newcommand{\conseggio}{$^\textrm{\small \yogh}$}
\newcommand{\universitasindonesia}{$^\textrm{\small \ae}$}
\newcommand{\supzurich}{$^\textrm{\scriptsize \" u}$}
\newcommand{\suppaair}{$^\textrm{\scriptsize $\pi$}$} 
\newcommand{\supnitap}{$^\textrm{\scriptsize $\alpha$}$} 
\newcommand{\suptyvandotru}{$^\textrm{\scriptsize $\theta$}$} 
\newcommand{\supwiki}{$^\textrm{\scriptsize \^{w}}$} 
\newcommand{\supmari}{$^\textrm{\scriptsize {\fontencoding{T2A}\selectfont м}}$} 
\newcommand{\capitalone}{$^\textrm{\scriptsize c}$} %

\newcommand{\suplij}{$_\textrm{\scriptsize lij}$}
\newcommand{\supdje}{$_\textrm{\scriptsize dje}$}
\newcommand{\supmos}{$_\textrm{\scriptsize mos}$}
\newcommand{\supyue}{$_\textrm{\scriptsize\begin{CJK*}{UTF8}{bsmi}{\CJKfamily{bkai}粵}\end{CJK*}}$}
\newcommand{\suptrp}{$_\textrm{\scriptsize trp}$}
\newcommand{\supid}{$_\textrm{\scriptsize id}$}

\newcommand{\supru}{$_\textrm{\scriptsize ru}$}
\newcommand{\supku}{$_\textrm{\scriptsize ku++}$}
\newcommand{\supnqo}{$_\textrm{\scriptsize nqo}$}
\newcommand{\suptum}{$_\textrm{\scriptsize tum}$}\newcommand{\supmhr}{$_\textrm{\scriptsize mhr}$}
\newcommand{\supsemicolon}{$_\textrm{\scriptsize ;}$}

\title{
\textsc{SMOL}: \\
Professionally translated parallel data \\  for 115 under-represented languages}

\author{Isaac Caswell\ugh{}$^*$
Elizabeth Nielsen\ugh{}$^*$
Jiaming Luo\ugh{}
Colin Cherry\ugh{} \\
    \bf{
Geza Kovacs\ugh{}
Hadar Shemtov\ugh{}
Partha Talukdar\ugh{}
Dinesh Tewari\ugh{}
    } \\
    \bf{
Baba Mamadi Diane\oho{}\supnqo{}
Djibrila Diane\oho{}\supnqo{}
Solo Farabado Cissé\oho{}\supnqo{}
    } \\
    \bf{
Koulako Moussa Doumbouya\otho{}\supnqo{}
Edoardo Ferrante\conseggio{}\suplij{}
Alessandro Guasoni \conseggio{}\suplij{}
    } \\
    \bf{
Christopher Homan\suppaair{}\supdje{}
Mamadou K. Keita\suppaair{}\supdje{}\supsemicolon{}\supmos{}
Sudhamoy DebBarma\supnitap{}\suptrp{}
Ali Kuzhuget\suptyvandotru{}\supru{}
    } \\
    \bf{
David Anugraha\otho{}\supid{}
Muhammad Ravi Shulthan Habibi\universitasindonesia{}\supid{}
Genta Indra Winata\capitalone{}\supid{}
    } \\
 \bf{ Anthony Munthali\supwiki{}\suptum{}
Sina Ahmadi \supzurich{}\supku{}
Andrei Chemyshev\supmari{}\supmhr{}
    } \\
   \bf{ Mingfei Lau\ugh{}\supyue{}  Jonathan Eng\ugh{}\supyue{}
 } \\
\ugh{}Google \{Research, Deepmind\}
\oho{}NKo USA INC
\otho{}Stanford University \\
\conseggio{}Conseggio pe-o patrimònio linguistico ligure
\universitasindonesia{}Universitas Indonesia
\supzurich{}University of Zurich \\
\suppaair{}Rochester Institute of Technology
\capitalone{}Capital One \suptyvandotru{}tyvan.ru \\ \supmari{}Mari Institute of Language, Literature and History\\[-1ex]
\noindent\rule{0.3\linewidth}{0.2pt} 
\\
\raisebox{0.5ex}{\supnqo{}\supsemicolon{}\suplij{}\supsemicolon{}\supdje{}\supsemicolon{}\supmos{}\supsemicolon{}\suptrp{}\supsemicolon{}\supru{}\supsemicolon{}\supid{}\supsemicolon{}\supku{}\supsemicolon{}\suptum{}\supsemicolon{}\supmhr{}\supsemicolon{}\supyue{}}
Led volunteer contributions for NKo, Ligurian, Zarma, Moor{\'e}, Kokborok, \\Russian, Indonesian, Hawrami+Gilaki+Sorani, Tumbuka, Meadow Mari, and Cantonese, respectively  \\[-1ex]
\noindent\rule{0.3\linewidth}{0.2pt} 
\\
  {\small {$^*$}\texttt{\{icaswell, eknielsen\}@google.com}} \\
}

\begin{document}
\maketitle
\begin{abstract}
We open-source \smol{} (\textit{Set of Maximal Overall Leverage}),\footnote{\raisebox{-0.15cm}{\includegraphics[height=0.5cm]{hf-logo.png}}~
\href{https://huggingface.co/datasets/google/smol}{\textbf{google/smol}}} a suite of
training data
to unlock machine translation for low-resource languages. \smol{} has been translated into \nsmolnonenglishlangs{} (and growing) under-resourced languages (\nsmolnonenglishps{} language pairs),\footnote{Experiments are mostly on a subset of 115 languages, before  volunteer translations of additional languages  finished. The paper title reflects this.} including many for which there exist no previous public resources, for a total of 6.1M translated tokens. \smol{} comprises two sub-datasets, each carefully chosen for maximum impact given its size:   \smolseal{}, a set of sentences chosen for broad unique token coverage, and \smoldog{}, a document-level resource focusing on a broad topic coverage.
They join the already released \gatitos{} for a trifecta of paragraph, sentence, and token-level content.
We demonstrate that using \smol{} to prompt or fine-tune Large Language Models yields robust \chrf{} improvements. In addition to translation, we provide factuality ratings and rationales for all documents in \smoldog{}, yielding the first factuality datasets for most of these languages.
\end{abstract}

\section{Introduction}
There exist no professionally-translated data for most of the world's 7000 or so languages, rendering tasks like Machine Translation near impossible. High-quality data is needed. However, it is not clear how best to use a limited budget for an expensive task like professional translation.
As shown by the \gatitos{} dataset \citep{Jones2023}, word-level translations provide large benefits to translation quality for low-resource languages at the lowest cost.
However, gains quickly saturate, as single tokens are not very expressive.
Sentence-level data is better for a model once token-level data saturates, but it has much more inherent redundancy; and document-level data is even more effective...and more redundant.

In this work, we release the \smol{} dataset, which provides professionally translated sentence- and document-level data for \nsmolnonenglishlangs{} LRLs (\nsmolnonenglishps{} language pairs). \smol{} contains two sub-datasets:


\begin{itemize}
    \item \textbf{\smolseal{}}: 863 English sentences covering 5.5k of the most common English tokens,\footnote{In this paper, `token' refers to typographic units as an approximation to words, not subword tokens from a model's vocabulary.}
    professionally translated into \nsmolsealnonenglishlangs{} languages.
    \item \textbf{\smoldog{}} \nsmoldogdocs{} English documents covering a wide range of topics, domains, and tokens, generated by an LLM and professionally translated into \nsmoldognonenglishlangs{} languages.

\end{itemize}

We demonstrate the utility of these data for finetuning and prompting LLMs for translation, and provide factuality annotations for all documents.

\section{Related work}

There are not many training datasets with human-translated data for Low-Resource Languages (LRLs), where we operationally define LRL as any language beyond the first 100 supported by most traditional crawls and MT providers (enumerated in Appendix section \ref{app:lrl-def}).

Tatoeba \citep{tatoeba} is probably the most multilingual, but it is made of volunteer contributions and of unclear quality.
The \gatitos{} dataset \cite{Jones2023} consists of a 4000-entry lexicon translated into 170 LRLs, but is only token-level. Most similar to the present work, \nllbseed{} is a high-quality, sentence-level training set of 6k sentences selected from English Wikipedia and professionally translated into 44 LRLs \cite{Nllb2022}.
There are also several professionally-translated evaluation sets, namely \floresone{} and \florestwo{} \cite{Goyal2022, Nllb2022}, and \ntrex{} \citep{ntrex}.

While highly multilingual, professionally translated training data is rare,
there is a growing number of bottom-up community data sources organized through research collectives like Masakhane \citep{masakhane}, Turkish Interlingua \citep{mirzakhalov2021large, mirzakhalov2021evaluating}, and GhanaNLP~\citep{ghananlp};
and conferences and workshops like AfricaNLP,
AmericasNLP~\citep{mager-etal-2021-findings} and ArabNLP.
These datasets are usually generated by researchers fluent in the languages, and are therefore especially high quality.
In addition to providing datasets, such efforts frequently also provide models and baselines, or even public interfaces, like the Khaya Translator Web App\footnote{\scriptsize{\url{https://ghananlp.org/project/translator-webapp/}}} by GhanaNLP for West African languages, and the lesan.ai\footnote{\scriptsize{\url{https://lesan.ai/translate}}} translation website for Ethiopian languages.

Participation is especially strong from the African continent, including corpora and models for pan-East-African languages \citep{babirye2022building}, languages from the Horn of Africa \citep{hornmt}, Ethiopian languages \citep{teferra-abate-etal-2018-parallel,gezmu2021extended}, Ugandan languages \citep{akera2022machine}, Emakhuwa \citep{felermino2021towards},  South-African languages \citep{eiselen-puttkammer-2014-developing}, Setswana and Sepedi \citep{marivate-etal-2020-investigating}, Yorùbá \citep{adelani-etal-2021-effect,adelani2021menyo},  Oshiwambo \citep{nekoto2022participatory}, Igbo \citep{ezeani2020igbo},
Zulu~\citep{rooweither_mabuya_2021_5035171},
Twi \citep{azunre2021english}, Gbe \citep{hacheme2021english2gbe}, Bambara \citep{tapo2021domain}, and Fon \citep{emezue-dossou-2020-ffr}. Outside of Africa, corpora have been created for languages of the Americas, including for four indigenous languages of Peru in \citet{bustamante-etal-2020-data}, the numerous results on the largely South- and Central American languages from the first AmericasNLP conference \citep{mager-etal-2021-findings}, and the Inuktitut language of Canada \citep{joanis-etal-2020-nunavut}. Datasets for lower-resourced languages of India have also sprung up, including the 13-language PMIndia \citep{haddow2020pmindia}, and datasets focused on languages of the Northeast like Mizo \citep{thihlum2020mizo}, Khasi \citep{laskar-etal-2021-enkhcorp1} and Assamese \citep{laskar-etal-2020-enascorp1}. Further West, PARME \citep{ahmadi2025parme} has provided some of the first human-translated content for Kurdish and Iranian languages. Finally, a variety of such datasets and models are available for public use on HuggingFace\footnote{\scriptsize{\url{https://huggingface.co/datasets?multilinguality=multilinguality:translation&task_categories=task_categories:translation}}} or Zenodo.\footnote{\scriptsize{\url{https://zenodo.org/communities/africanlp/}}}

In addition to professionally translated data, there are also several web-crawled datasets for LRLs, including \textsc{MadLad} \citep{Kudugunta2023}, OSCAR~\citep{oscar}, Glot500-C~\citep{imanigooghari-etal-2023-glot500}, NLLB \citep{Nllb2022}, and the Bloom library \citep{leong-etal-2022-bloom}.

\section{Text Selection}
\label{sec:methods}
Translation requires significant investment and can't be easily re-done, so great care needs be put into carefully choosing sentences to translate.
For both sub-datasets \smoldog{} and \smolseal{},
selection or generation of source text is done in English. Selecting only English has clear biases, but also has advantages---most notably, for N languages, it requires N times less work to quality control. Future work should consider focusing on non-English sources.



\subsection{\smolseal{}: Token Set Cover}
\label{sec:tsc}
Our basic motivation for creating \smolseal{} was to help models overcome vocabulary issues, which are common for the lowest-resource languages \citep{alligators,bapna2022building}.
Therefore, we frame this as a set-cover problem, and pick the smallest set of sentences (from  CommonCrawl\footnote{https://commoncrawl.org/ we use all available snapshots as of August 20, 2022}) that covers the largest set of target tokens. The tokens we chose to cover
(the \textit{target set})
were the English side of \gatitos{}, as well as the most common 2,500 tokens from an English web crawl. Set cover is NP-hard, so we approximate it with a greedy algorithm that iteratively picks the sentence with the highest \textit{coverage percent}, defined as the percentage of its tokens that are in the target set.

\paragraph{Preliminary work on Token Set-Cover}
\label{sec:internship}

\begin{table}[]
    \centering
    \small
    \begin{tabular}{lc}
        \toprule
        \textbf{Method} & \textbf{ChrF} \\
        \midrule
        Random &  30.5\\
        Token set-cover & \textbf{31.7}\\
        N-gram DWD & 30.0\\
        Embedding DWD & 27.5 \\
        \bottomrule
    \end{tabular}
    \caption{
        Held-out \chrf{} for data selection approaches
    }
    \label{tab:selectioneval}
\end{table}

To evaluate the token set-cover approach, we started by selecting data from existing web-scraped parallel data.
We pretrain a multilingual Neural Machine Translation (NMT) model on parallel data from 294 language pairs from \madlad{} \cite{Kudugunta2023},
with nine languages held out to simulate LRLs.
We fine-tune this model on sets of existing parallel data in each of the held-out languages, and evaluate on \florestwo{}. Details
on the experimental set-up
in Appendix \ref{app:selectioneval}.

In addition to Greedy Token Set-Cover, we explore two methods that balance data diversity and data quality.
First, we implement \citet{ambati2011}'s `density-weighted diversity' (DWD) metric, which is an $n$-gram based metric for diversity and quality.
Second, we implement an embedding-based version of DWD, which takes the weighted harmonic mean of perplexity under the Palm 2 model \cite{anil2023} (proxy for quality), and embedding distance on mBERT sentence embeddings (proxy for diversity).
We apply both methods to the English side of the parallel data only, to simulate the case where we don't yet have LRL translations.
As a baseline, we randomly select sentences.

Table \ref{tab:selectioneval} shows results after finetuning. Greedy token set-cover performs the best, with diversity-based metrics actively hurting performance.

\paragraph{Researcher in the Loop (RITL)}
\label{ritl}

Despite its success in the ablation, Greedy Token Set Cover had several problems when we scaled it to select from among all the English sentences of CommonCrawl. Firstly, it is maximized by honeypots, or nonsense strings dense in content words (Appendix Table \ref{tab:badgreedy});
and secondly, it biases towards short sentences,
causing length distribution artifacts.

These problems are not easy to solve with heuristics---for example, if you disqualify lists with commas you'll get ones with spaces, if you require sentences to have some function words or token-length diversity, you'll get other sorts of garbled sentences, and so on. However, a dataset like \smol{} is small enough to manually inspect. Therefore we develop
\textit{Researcher in the Loop Greedy Set-Cover} (Algorithm \ref{alg:ritlgsc}), where the domain expert (the researcher) can look at and edit each individual sentence.\footnote{This work was conducted before the advent of LLMs. Today, this could be simplified using LLMs as autoraters.}
The result of this process is \smolseal{}, a set which uses 863 sentences to cover 5519 unique tokens. Qualitatively, \smolseal{} consists of complex sentences with wide vocabulary coverage; quantitative metrics are explored in Appendix \ref{app:smolsealstats}.

\begin{algorithm}
\small
\caption{Researcher in Loop Greedy Set Cover}
\begin{algorithmic}
\State Res $\gets$ ...  \Comment{Sentence reservoir, e.g. CommonCrawl}
\State Toks $\gets$ ...  \Comment{Tokens to Cover, e.g. GATITOS}
\State Cov $\gets$ \{\} \Comment{Set-cover, aka output of this algorithm}

\While{not ToCover.empty()}
 \State batch $\gets$ TopScoringSentences(Res, Toks)
 \State chosen $\gets$ ResearchersChoice(batch)
 \State chosen $\gets$ LetResearcherEdit(chosen)
  \State Cov.add(chosen)
  \State RemoveCoveredToks(Toks , chosen)
 \State Res $\gets$ LetResearcherDiscardSentences(Res)
  \State Res.remove(chosen)
\EndWhile
\State return Cov

\end{algorithmic}
\label{alg:ritlgsc}
\end{algorithm}

\label{smolseal:methods}

\subsection{\smoldog{}: LLMs with prompt mesh}
\label{smoldog:methods}
\smoldog{}
follows a different and complementary approach. Whereas \smolseal{} consists of a small set of \textit{sentences} that are \textit{selected} from natural text, are \textit{complex}, and cover many \textit{tokens}; \smoldog{} instead consists of \textit{documents} that are \textit{generated} and are  \textit{simpler},
but cover many \textit{topics}. It should be noted that the token-coverage approach described above failed resoundingly for longer documents, as the prevalence of the honeypots was magnified.

To generate \smoldog{}, we used a collection of templates to create a few thousand diverse prompts with a wide range of topics, domains, words, tenses, grammatical cases, and registers (e.g. formal, informal).
Appendix \ref{app:smoldogprompts} gives details and examples.

\paragraph{Corpus Diversity Ranking for \smoldog{}}
\label{sec:corpusdiversity}
Document-by-document evaluation as described above does not help one understand \textit{corpus diversity}---for example, if an almost identical document appears twice, only one of them should be included.
Therefore, we rank all candidates by how much new information they add to the corpus, by
iteratively finding the document contributing the least new information and removing it, thus ranking all documents.
Our criterion for ``new information'' was the average character 9-gram Inverse Document Frequency (IDF) score of a document---in other words, how rare its substrings were across all of the documents in the pool so far. To down-weight internally repetitive documents, we substracted the fourth moment \bread{} score \citep{bread}.

\paragraph{Language Tiers for \smoldog{}}
We wanted to translate more data for languages with more speakers. We break the languages into the five different groups, each with a larger subset of the generated documents. Each tier contains translations of the top N documents as ranked by corpus diversity.
These can be seen in Appendix Table \ref{tab:smoldogtiers}.

\paragraph{Non-English-centric translations}
For \smoldog{}, we additionally collected data for four non-English-centric language pairs, from each of the East African languages of Amharic (\texttt{am}) and Swahili (\texttt{sw}) to each of regionally relevant languages Standard Arabic (\texttt{ar}) and Mandarin Chinese (\texttt{zh}). Including the reversed versions of these, this yields 8 total language pairs. Because of the difficulty of generating good source material in these languages, we used the existing \smoldog{} translations to Swahili/Amharic as the source text. However, due to the lack of appropriate evaluation sets, it is hard to know the value-add of this data over datasets pivoted through English.

\section{Data Collection and Verification}
Several languages are contributed by volunteers; they are listed as co-authors.\footnote{Community contributions of translations or corrections are welcome; please reach out to the authors or join the TUSL Discord.} For the other languages, the translation provider we contracted has worked with us for many years, and has a pre-existing relationship with professional translators for all languages in the \smol{} datasets. The translators are paid a fair wage, and their identities are contractually kept anonymous to us.
We checked the delivery for duplicate translations, anomalous source/target length ratios, and similarity with Google Translate outputs.
Very few languages were flagged this way. Following this, we ran \textsc{FunLangID} \citep{funlangid} on all segments and found no issues. 
 Manual inspection turned up several issues with nonunicode fonts (e.g. {\^ o} for \openo ) for West African languages, and nonstandard orthography for Santali; these issues were then fixed. The choice of script, orthography and translation variety was challenging for many communities, including Kurdish, Zaza-Gorani and Gilaki languages, all of which have more than one orthography and lack a standard variety.

The largest missing check is for fluency, which is hard to measure without trusted native speakers \textit{outside of the translation agency}, or trusted LLMs; neither of which exist for all \smol{} languages.




\begin{table*}[t]
\centering
\small
\begin{tabular}{@{}lrrrrrrr@{}}
\toprule
 & \multicolumn{4}{c}{Total Dataset} & \multicolumn{3}{c}{Per Language Pair (LP)} \\
\cmidrule(lr){2-5} \cmidrule(lr){6-8}
Set & \# languages & Examples & Tokens & Characters & Examples & Tokens & Characters \\
\midrule

\gatitos{} & 178 & 714k & 807k & 4.8M & 4.0k & 4.5k & 27k \\
\smolseal{} & 88 & 76k & 1.1M & 6.6M & 863 & 12k & 75k \\
\smoldog{} & 106 & 27k & 5.1M & 29M & 256 & 48k & 270k \\
\midrule
\textsc{Smol}\{\textsc{Sent+Doc}\} & 125 & 103k & 6.2M & 35M & 824 & 50k & 282k \\

\bottomrule
\end{tabular}
\caption{Statistics for the whole data set (left bloc) and per language-pair (LP) (right bloc) on the two \smol{} datasets and their predecessor \gatitos{}
in number of examples, tokens, and characters. The \texttt{\# languages} column counts translated languages only, not the source languages of English, Swahili, and Amharic.
}
\label{tab:data}
\end{table*}

\begin{table*}[htp]
\centering
\small
\begin{tabular}{@{}lrrrrrrrr@{}}
\toprule
LP subset $\rightarrow$ & \multicolumn{2}{c}{\textsc{Smol-Sent} {\scriptsize (80 LP)}} & \multicolumn{2}{c}{\textsc{Smol-Doc} {\scriptsize (73 LP)}} & \multicolumn{2}{c}{Intersect {\scriptsize (38 LP)}} & \multicolumn{2}{c}{\textsc{Hard} {\scriptsize (32 LP)}}\\
\cmidrule(lr){2-3} \cmidrule(lr){4-5} \cmidrule(lr){6-7} \cmidrule(lr){8-9}
Model $\downarrow$ & 0-shot & 10-shot & 0-shot & 10-shot & 0-shot & 10-shot & 0-shot & 10-shot \\
\midrule
\textsc{G. Translate}      & -      & -      & -      & -      & \textbf{43.2} & -      & -      & -      \\
\textsc{NLLB-54b}          & -      & -      & -      & -      & 40.0          & -      & -      & -      \\
\textsc{Claude 3.5 Son.}   & 37.5   & \textbf{39.7} & 38.3   & \textbf{40.9} & 41.0          & 42.8   & 30.0   & \textbf{33.5} \\
\textsc{GPT-4o}            & 29.9   & 34.1   & 31.8   & 36.3   & 35.4          & 38.5   & 15.9   & 23.7   \\
\textsc{Gemini 2.0 Pro}    & \textbf{38.9} & 38.9   & \textbf{39.9} & 40.3   & 42.6          & 42.2   & \textbf{31.4} & 31.7   \\
\midrule
\textsc{Gemini-2.0 Flash}  & 35.6   & 38.4   & 36.9   & 39.7   & 40.2          & 41.4   & 26.3   & 30.4   \\
\textsc{+ \smolseal{}}     & 38.3   & 38.3   & 38.8   & 38.8   & 40.6          & 40.6   & 32.5   & 32.6   \\
\textsc{+ \smoldog{}}      & 35.3   & 35.4   & 39.5   & 39.5   & 41.2          & 41.2   & 31.8   & 31.8   \\
\textsc{+ \textsc{Both}}   & 38.9   & 38.9   & 40.5   & 40.5   & 41.8          & 41.8   & 33.4   & 33.4   \\
\textsc{+ \textsc{Both+G}} & \textbf{39.4} & \textbf{39.3} & \textbf{41.0} & \textbf{40.9} & \textbf{42.1} & \textbf{42.2} & \textbf{33.9} & \textbf{33.9} \\
\midrule
\textsc{$\Delta_{FT}$}     & +3.8   & +0.9   & +4.1   & +1.2   & +1.9          & +0.8   & +7.6   & +3.5   \\
\bottomrule
\end{tabular}
\caption{
    Finetuning Gemini 2.0 Flash on \smol{} for four subsets of language pairs. The first two columns show LPs in \smolseal{} and those in \smoldog{}, to show the different effects of each split. The third shows those in both \smol{} datasets AND the closed domain NMT models, for an even comparison to NMT models. Finally, the \textsc{Hard} column shows LPs in both \smol{} splits but NOT in Google Translate, or not closely related to a language in Google Translate, to approximate the especially hard languages to learn.
}
\label{tab:main-results}
\end{table*}

\section{Finetuning and In-Context Learning}

We use fine-tuning and ICL as tools to demonstrate the value of the \smol{} dataset. As this is a data paper, these experiments are motivated by the maxim \textit{``what could any researcher simply train with public APIs?''} More involved techniques, e.g. Reinforcement-Learning-based approaches, will likely lead to stronger results.

\subsection{Evaluation}
 Since so many language pairs are covered, we evaluate on a combination of all available evaluation sets, namely \florestwo{} \citep{Nllb2022}, \ntrex{} \cite{Federmann2022,Barrault2019},
and an in-house eval set. 
Since no reliable embedding models exist for these languages, trained metrics are not an option, so we use \chrf{}~\cite{popovic2015chrf} as implemented in SacreBleu \citep{post2018clarityreportingbleuscores}\footnote{signature: \scriptsize{\texttt{case.mixed+numchars.6+numrefs.1\\
+space.False+tok.none+version.1.3.0}}} with NFKC unicode normalization as our metric. For ten-shot decoding, exemplars were selected from both sub-datasets of \smol{} using \chrf{}-counterweighted RAG (Appendix \ref{sec:scrag}).

\subsection{Finetuning Setup and Results}
We finetuned Gemini 2.0 Flash for 40 epochs on \smoldog{}, \smolseal{}, a combination of the two (\smolboth{}), and their combination plus \gatitos{} (\smolbothg{}). To simplify finetuning, we split \smoldog{} into sentence pairs (\smoldogsent{}).

Results can be seen in Table \ref{tab:main-results}. Finetuning on \smolseal{} gives an average gain of +2.7 \chrf{} points, and \smoldogsent{} gives +2.6 \chrf{} points on its languages. Concatenating the two training datasets leads to a gain of +3.3 to +3.6 \chrf{} points, and adding in \gatitos{} bumps it up to +3.8 to +4.1 \chrf{} points, passing all baselines except Google Translate.
The 10-shot RAG results on the un-tuned model are very close to the finetuned 0-shot results, and the finetuned models show no benefit from multi-shot decoding, suggesting that these are two different ways of giving the same information---inference-time versus training time.
The 10-shot random results (not included in table) were much lower.

Gains were highest on languages that are not related to mid- or high-resource languages, and lowest on dialects close to major languages. As a heuristic to measure this, we exclude languages that are on Google Translate or closely related to languages on it (Appendix \ref{app:relatedlangs}). The average gain on these languages jumps to +7.6 \chrf{}.

\begin{figure}[t!]
\begin{center}
\includegraphics[scale=0.36]{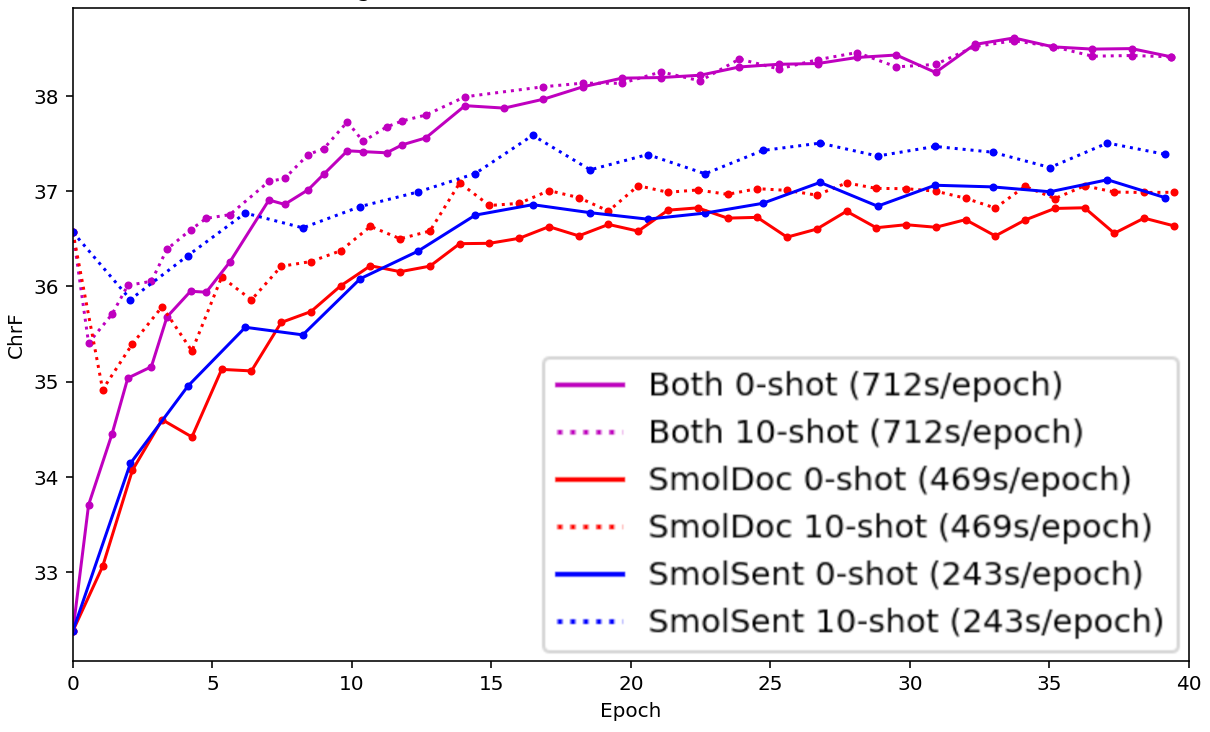}
\caption{Training curves (\chrf{}) for finetuned models on a subset of 37 \enxx{} language pairs, trained on \smoldog{}, \smolseal{}, and their combination \smolboth{}.
}
\label{fig:epochs}
\end{center}
\end{figure}

Figure \ref{fig:epochs} shows the learning curve on a development subset of 37 languages.
Although it may be surprising that so many epochs are needed before convergence, we found that further increasing learning rate led to overfitting. The sharp drop near the beginning suggests a domain mismatch between pretraining and finetuning, and suggests that the same data could be used much more effectively with a better training set-up than explored here.

\subsection{The Problem with \xxen{} training}
\label{sec:mwp}
Our initial experiments used all data for both \enxx{} and \xxen{}.
However, the models lost performance on all tasks.
The root cause turned out to be the multiway-parallel data with English targets.
LLMs are especially susceptible to repetition in data \citep{Lee2022}, and with 115 language pairs, for every one epoch over the data, the model saw about 115 epochs for each individual target sentence. Therefore, it wildly overfit and lost performance on all language pairs.
Mitigating such overfitting is an important research direction to pursue, since many promising datasets are multiway parallel, e.g. \floresone{} \citep{Goyal2022}, \florestwo{} \citep{Nllb2022}, \ntrex{} \cite{Federmann2022,Barrault2019}, and others.
However, this is out of scope for the present paper,
so we restrict our experiments to \enxx{}.

Seeing the same \textit{source} many times likely also has deleterious effects and should also be studied; but these effects, if they exist, are small enough that we were still able to see net gains.

\section{Factuality Review}

\begin{figure}[t!]
\begin{center}
\includegraphics[scale=0.19]{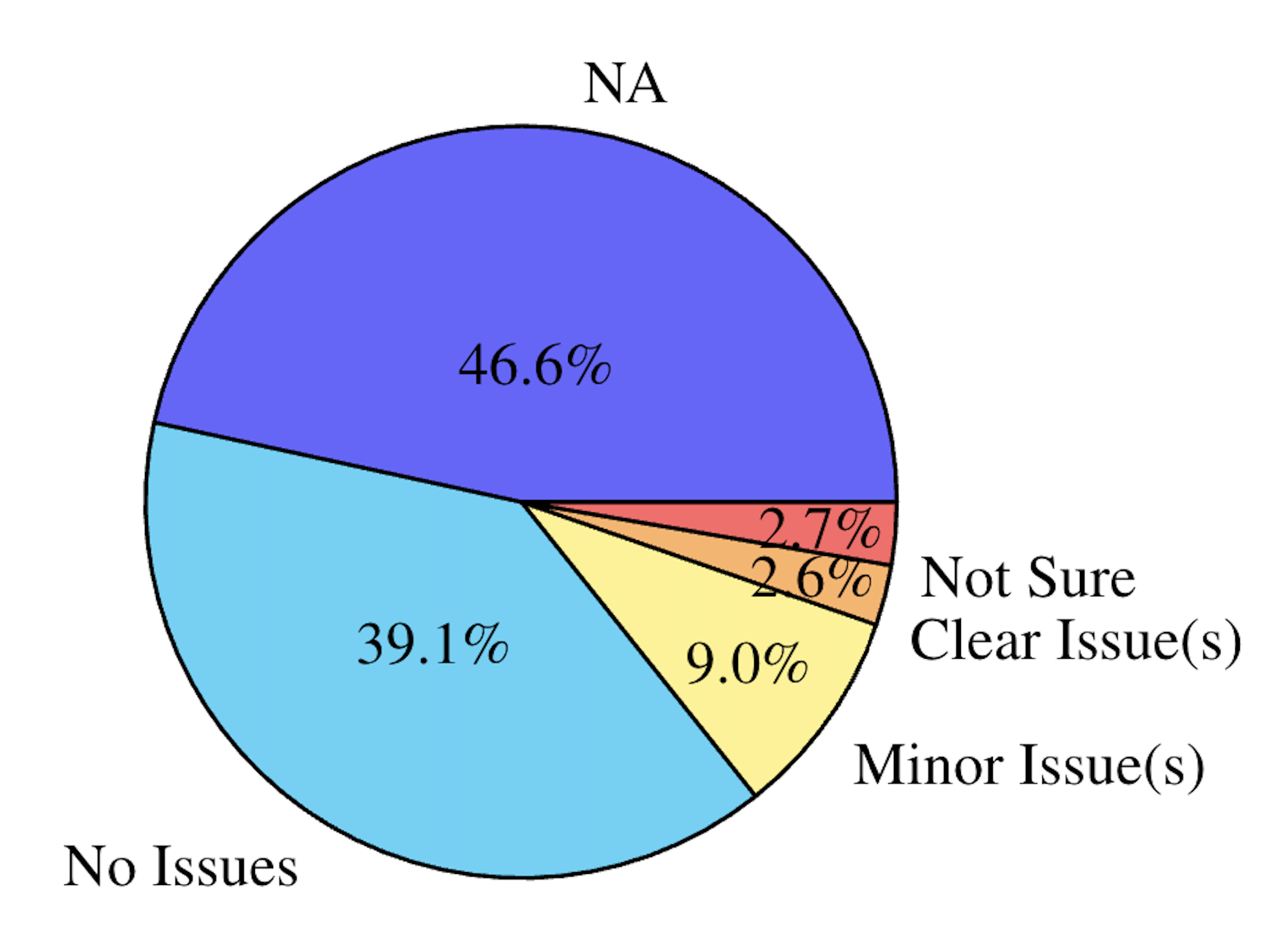}
\caption{\mbox{\smoldog{} factuality ratings}.
}
\label{fig:fact-pie}
\end{center}
\end{figure}

Since \smoldog{} contains LLM-generated sources, they contain some factual inaccuracies.
We therefore do a full human audit and assign factuality codes to each document.
Each of the \nsmoldogdocs{} documents was rated by three raters. Each rating is accompanied by a detailed explanation, including sources cited. Inter-annotator agreement was high, with Cohen's $\kappa$ between each pair of raters between 0.82-0.87. The error code distribution can be seen in Figure \ref{fig:fact-pie}. The rubric is presented in Table \ref{tab:fact-rubrix}.

\begin{table*}[tbp]
\centering
\small
\begin{tabular}{l|p{13cm}}
\hline
\hline
Rating & 	Definition of Rating \\
\hline
N/A  & 	True/False does not apply here. Most stories, dialogues, or fictional works would be considered N/A, unless they are promoting a falsehood about the real world. \\
Not Sure & 	Claims are made that may not be true, but you aren't sure. Choose this if it would take over 10 minutes to verify the factuality of the claim. \\
No Issues & 	All claims are factual and accurate. (Out-of-date is fine, e.g. ``Barack Obama is the US President'') \\
Minor Issue(s) & 	There are small inaccuracies. E.g., it may be broadly correct but frame something in a misleading way. \\
Clear Issue(s) & 	There are clear mistakes in factuality. \\
\hline
\end{tabular}
\caption{Factuality Rubric}
\label{tab:fact-rubrix}
\end{table*}



All ratings and rationales are made available. In addition, each datum in \smoldog{} is given with a simple \texttt{factuality} annotation, which has the value \texttt{has\_errors} if any one of the ratings was any of \texttt{Minor Issues} or \texttt{Clear Issues}, and \texttt{ok} otherwise.
For some use-cases, like question-answering, practitioners may want to filter out nonfactual data; for others, like translation, one may not be troubled by factual errors. In addition to filtering, this also provides the first factuality dataset for most of these languages.

\section{Conclusion}
We have open-sourced the \smol{} dataset, a professionally-translated dataset covering \nsmolnonenglishlangs{} (and growing!) low resource languages and targeting the tasks of translation and factuality. It comprises \smoldog{} and \smolseal{}, two training datasets with the complementary strengths of sentence selection (complex, and high token coverage) and document generation (contextual, varied domains, simpler sentences) respectively.
We demonstrate that finetuning Gemini 2.0 Flash on these yields to substantial improvements in translation quality.
\smol{} joins a growing body of resources to support underserved languages in the age of AI.

\section{Limitations}
The \smol{} data would benefit from a more thorough review, audit, and correction from community members outside of the translators who created it.
Future work on \smol{}-like datasets should also focus on non-English source text that is not only maximally authentic in the given language, but also covers the topics and concepts most relevant to those languages. This approach is more difficult and would require significant work and review to do correctly.
Finally, more research is needed to understand and prevent the overfitting that comes with multi-way parallel data.



\newpage
\bibliography{anthology}
\newpage
\appendix
\counterwithin{figure}{section}
\counterwithin{table}{section}

\section{Operational Definition of LRL}
\label{app:lrl-def}
In this paper, we operationally define LRL as any language beyond the first 100 supported by most traditional crawls and MT providers. Since there is some variation in which languages exactly this is, we concretize it as the set of 104 languages supported in Google Translate prior to 2020. These are the languages for which launchable quality was possible before LLM-type models like M4 \citep{arivazhagan2019massivelymultilingualneuralmachine, bapna2022building} and PaLM \citep{anil2023} came on these scene. It is also worth noting that since these languages were on the product for much longer, they have much more machine-translated content online from services that used Google Translate for internationalization. These 104 languages are: \texttt{af am ar az be bg bn bs ca ceb co cs cy da de el en eo es et eu fa fi fil fr fy ga gd gl gu ha haw hi hmn hr ht hu hy id ig is it he ja jv ka kk km kn ko ku ky la lb lo lt lv mg mi mk ml mn mr ms mt my ne nl no ny pa pl ps pt ro ru sd si sk sl sm sn so sq sr st su sv sw ta te tg th tr uk ur uz vi xh yi yo zh zh-Hant zu}.

Rightly speaking, the languages outside of this set might better be termed "Very Low-Resource" instead of just "Low Resource", since the 104 languages above do include languages like Hawaiian, Javanese, Yiddish, and Hmong, which are by no stretch of the imagination high-resource.  We will leave more rigorous definitions to future work.

\section{\smolseal{} details}
\label{app:smolseal}

\subsection{Evaluating the \smolseal{} selection process}
\label{app:selectioneval}

In Section \ref{sec:internship}, we describe experiments used to validate the selection process for \smolseal{}.
We train a backbone MT system is pretrained on the \madlad{} dataset \citep{Kudugunta2023}.
The following languages are held out of the training data to be used for fine-tuning experiments: Catalan, Icelandic, Marathi, Turkish, Maltese, Xhosa, Tamil, Basque, and Tajik.
The model itself is a 1B parameter encoder-decoder Transformer and is trained from scratch on the \madlad{} data.
Each of the candidate data selection methods is used to select data from the held-out languages, and then each candidate set is used in turn to finetune the backbone model. The results of this finetune step are reported in Table \ref{tab:selectioneval}, where the set-cover approach is shown to be most effective.

\subsection{Notes on Researcher in the Loop}
\label{app:ritl}
Researcher In the Loop extends the greedy set cover approach thusly: rather than always picking the highest-scoring sentence, we iteratively show the researcher a batch of the 20 highest scoring sentences according to several scores, and let the researcher pick and optionally edit each sentence at each iteration. At each iteration, the researcher may also remove any number of this batch's sentences from the reservoir.
Allowing the researcher to see and edit the sentences allows ensures that the sentences are of high-quality. To deal with the length bias issue, we showed not only sentences that maximize coverage percent, but also that maximize heuristics that weighted the coverage with the number of new tokens hit, like \texttt{ log(coverage\_percent)*n\_hits }.

As described in the paper text, this approach is designed to combat issues such as honeypot sentences. Example Honeypot sentences can be seen in Table \ref{tab:badgreedy}

\begin{table}[h]
\scriptsize
\centering
\begin{tabular}{p{0.45\textwidth}} 
\toprule
\textbf{Sentence} \\
\midrule
Individual determine can get prolonged, reduce along with attractive. \\
\midrule
Sell hand mood situation connect proper decision today spread true. \\
\midrule
Demand indeed off forget act special well treat sometimes notice. \\
\midrule
Agree board book oh trust by attractive supply deal together. \\
\midrule
Picture exactly could ability impact advance then same admire across. \\
\midrule
One physically courage both information language issue laugh common. \\
\bottomrule
\end{tabular}
\caption{Honeypot Sentences for Greedy Selection: CommonCrawl has many sentences packed with content words but with no clear semantics or grammar.}
\label{tab:badgreedy}
\end{table}

\subsection{Corpus statistics on \smolseal{}}
\label{app:smolsealstats}

To measure ``Bang for our Buck'' we define the \textit{excess-token ratio} $\xi$ as the number distinct tokens in the set cover divided by the number of target tokens, and use it along with the coverage percent to understand the \smolseal{} dataset. Table \ref{tab:smolseal} compares corpus statistics of \smolseal{} to four other corpora. \texttt{sametoks} picks a random set of sentences from CommonCrawl until it has the same number of tokens as \smolseal{}; this only covers 50\% of the target tokens and has an excess-token-ratio $\xi$ ratio of 3.3, much worse than \smolseal{}'s value of 2.3. The \texttt{samecov} baseline randomly picks common-crawl sentences until it has the same token coverage as \smolseal{}, which necessitates a 67x larger set of sentences, and a correspondingly bloated excess token coverage ratio of 23.1. As a further reference we compare tiers 1 (largest) and 5 (smallest; comparative size to \smolseal{}) of \smoldog{}. As expected of machine-generated text, they have a worse $\xi$ value, corresponding to a narrower spectrum of vocabulary used.

\begin{table*}[t]
\centering
\small
\begin{tabular}{llllrl}
\hline
\hline
set & N sent & toks  & types &   $\xi (\downarrow)$ & cov\%$(\uparrow)$     \\
\hline
\smolseal{} & 863 & 12k & 5.5k & 2.3 & 99.6\% \\
{\tt sametoks} & 863 & 12k & 3.8k & 3.3 & 50.4\% \\
{\tt samecov} & 57578 & 877k & 38k & 23.1 & 99.6\% \\
\hdashline
\smoldog{}-\textsc{t}1 & 6979 & 108k & 8.8k & 12.3 & 80.5\% \\
\smoldog{}-\textsc{t}5 & 820 & 12k & 2.8k & 4.3 & 40.2\% \\

\hline
\end{tabular}
\caption{Corpus statistics of \smolseal{}, random selections of sentences from CommonCrawl, and tiers 1 and 5 of \smoldog{}.
}
\label{tab:smolseal}
\end{table*}

\section{\smoldog Details}

\subsection{\smoldog{} data tier details}
Per-tier statistics on the \smoldog{} dataset can be seen in Table \ref{tab:smoldogtiers}.

\begin{table*}[t]
\centering
\small
\begin{tabular}{l|rrrr:rrr}
\hline
\hline
set & N langs & ex  & tok & char  & ex/LP  & tok/LP & char/LP     \\
\hline
\smoldog{}-t1 & 5 & 2.9k & 537k & 3.0M & 584 & 107k & 604k \\
\smoldog{}-t2 & 31 & 14k & 2.7M & 15M & 450 & 88k & 496k \\
\smoldog{}-t3 & 24 & 6.7k & 1.2M & 6.8M & 280 & 50k & 281k \\
\smoldog{}-t4 & 8 & 1.0k & 184k & 1.0M & 126 & 23k & 128k \\
\smoldog{}-t5 & 38 & 2.5k & 448k & 2.5M & 66 & 12k & 65k \\
\hline
\end{tabular}
\caption{Statistics on the languages in the individual tiers of \smoldog{}.}
\label{tab:smoldogtiers}
\end{table*}

\subsection{Details on \smoldog{} prompt creation}
\label{app:smoldogprompts}

To avoid biases from overly tempted prompts, we put in significant effort to make sure the prompts were all very different. Each prompt drew at random from the following elements:

\begin{itemize}
    \item random selection of English words to use in the response
    \item one of 600 manually created topics, e.g. ``volcanic eruptions'' or "A special tree''
    \item one of 50 tone/tense categories, e.g. ``Please use the subjunctive mood.'', ``Use an effusive tone.''
    \item A style prompt, e.g. ``You are the author R.K. Narayan.'' or ``You are a mother talking to her son.''
    \item A text modality, e.g. story/dialogue/essay
\end{itemize}

In addition to this, we added a few more sources of prompts:

\begin{itemize}
    \item Prompts based on urls, meant to simulate different web domains, like Wikipedia and reddit.
    \item Prompts based on continuing the sentences from \smolseal{}
    \item Prompts based on current events, history, and daily life in different countries
    \item Special effort was made to include dialogues (to get more spoken register) and recipes (unique domain that may also be important to translate).
\end{itemize}

For each prompt, we generated 8 responses (T=0.7). These were ranked by their simple token density (unique tokens over total tokens), and the top two were chosen for consideration. Using the Researcher-in-the-loop mentality (``measure twice cut once!''), we went over 1000+ responses by hand and scored/edited them. This was mainly to filter out questionable or boring responses. A typical paragraph scored as 0 would be LLM-speak like \textit{``X is a complex and multifaceted problem with no easy solution. Here are some suggestions. Keep in mind that there is no one-size-fits all solution, and ultimately, the choice is up to you. [...]''}.

Example prompts can be seen in Table \ref{tab:exampleprompts}.

\begin{table*}[t]
\centering
\footnotesize
\begin{tabular}{p{12cm}}
\hline
\hline
\textbf{Example Prompts} \\
\hline
You are Ernest Hemingway. Write a dialogue about road rage. Use a didactic tone.  \\
\midrule
Write a 1-paragraph story concerning an Irish wake.  \\
\midrule
You are a teenager talking to his friend. Please carefully craft a 1-paragraph bit about an engineer who subsists off coffee. Try to include the words ``confirmed'', ``move'' and ``above''. \\
\midrule
Give a typical, yet interesting, example of something you would find on reddit. \\
\midrule
Please write a few paragraphs about challenges facing Ethiopia. \\
\midrule
Please write a long passage starting with `Mum and Dad pause their debate when we hear this creepy clacking that sounds like hail falling.' \\
\midrule
Write a recipe for baking an almond cake. \\
\hline
\end{tabular}
\caption{Representative sample of prompts use to generate the documents for \smoldog{}}
\label{tab:exampleprompts}
\end{table*}

\subsection{\smoldog{} Errata}
\label{app:smoldogerrata}
\paragraph{Orthographies} Several languages use irregular orthographies. Most notable is Moor{\'e}/Mossi (\texttt{mos}), where different translators have used a variety of different conventions. After soliciting community feedback, we plan to release standardized versions to the data.

\paragraph{document selection} When collecting the data for \smoldog{} for the Indian languages, we mistakenly included a variety of documents that fell below the corpus diversity threshold described in Section \ref{smoldog:methods}.

\section{N-shot: \chrf{}-counterweighted RAG}
\label{sec:scrag}
To have a strong baseline for N-shot results, we adopt a RAG-based approach that resembles the greedy set-cover algorithm. For each sentence in the eval set, we want the best coverage of the source sentence $n$-grams as possible, with the least redundancy among exemplars. Therefore, we iteratively choose the exemplar whose source side has the minimum \chrf{} to the eval source. However, when counting the true positives in the \chrf{} calculation, we weight the count of each ngram $n_i$ by $(1+c_i)^{-\alpha}$, where $c_i \in [0, \infty]$ is the number of times $n_i$ has been seen among the exemplars chosen so far, and $\alpha$ is a parameter to control how close this algorithm is to ngram set-cover. We use $\alpha=2$.The set of exemplars we choose from is the concatenation of \smolseal{} and \smoldogsent{}.


\section{Prompts for Decoding}
\label{app:inferenceprompts}

For 0-shot prompting, we used the following, fairly wordy prompt, the SL and TL standing for the source and target language name, respectively:

\begin{verbatim}
You are an expert translator. I am
going to give you some example
pairs of text snippets where the
first is in ${SL} and
the second is a translation of
the first snippet into
${TL}. The sentences
will be written
${SL}: <first sentence>
${TL}: <translated first sentence>
After the example pairs, I am
going to provide another sentence
in ${SL} and I want
you to translate it into
${TL}. Give only the
translation, and no extra
commentary, formatting, or
chattiness. Translate the text
from ${SL} to ${TL}.
\end{verbatim}

For finetuned models, there is no need for such a wordy prompt, and indeed it only risks overfitting. Therefore, we used the following minimalist prompt:

\begin{verbatim}
Translate from ${SL} to ${TL}:
\end{verbatim}

\section{Volunteer contributions}
A few languages have extra details that need to be called out here.

\subsection{Translations for Cantonese}
A volunteer team of Cantonese speakers at Google pulled together to translate the maximal set of \smol{} text. Mingfei Lau and Jonathan Eng were the main leaders of this effort, and the contributors to translation and post-editing were (alphabetically): Tsz Yan Au, Emily Awesome, Jason Chan, Siu Man Chan, Vicky Chan, Yiwang Chen, Kinton Cheung, Mingo Choi, Andy Chow, Ashley Chow, Olivia Chow, Daniel (Ying Wai) Fan, Thomas Fung, Vikki Ha, Joshua Kwong, Liam Lee Pong Lam, Jonas Lau, Ying Tung (Grace) Law, Crystal Lee, Aki Leung, Derek Leung, Jackie Leung, Thomas Leung, Mu Li, Alicia Liu, Malena Loosli, Chui McConnell, Ken Ng, Nicholas Ng, Tonia Shen, Helen Shum, Franky Sze, Eric Tang, Tommy Tse, Daniel Wong, Danny Wong, Maggie Wong, Pinki Wong, Jeffrey Yu, Shanelle Yu, Shing Fung Yue, Miranda Zhang, and Willis Zhang.

\subsection{Translations for NKo}
The initial delivery for the NKo language (\texttt{nqo}) had a wide variety of errors. We reached out to the authors from \citet{doumbouya2023machinetranslationnkotools}, who did a complete re-translation of the text.

\subsection{Translations for Zazaki, Hawrami, and Gilaki}

Sina Ahmadi gratefully acknowledges support from the UZH Postdoc Grant (reference number 269093).


\subsection{Translations for Zarma}
\paragraph{Annotation Pipeline}
The Zarma translation process of SMOL---all the subsets---was done through a combination of
automatic and human in the loop methods.
We leveraged some existing tools that our team developed to speed up the annotation process.
We first used a baseline bidirectional model that we developed to produce initial translation of
the samples. These machine translated samples were then passed through our Zarma
grammatical error correction model. This model was built by pre-training gemma-2-9b on Zarma
data and fine tuning the checkpoint on grammar error correction data set using Direct
Preference Optimization (DPO) settings. The outputs from this stage---both languages side by
side---were then given to our team of annotators for review.

The annotators were given some guidelines---in addition to the general guidelines from \smol{}---for the annotations. These guidelines include:
\begin{itemize}
\item \textbf{Word adaptation:} rules for handling technical terms, proper nouns, and domain-specific
vocabulary that might not have direct equivalents in Zarma. E.g: all the scientific/technical words
remain unchanged; and words that have known french-ized equivalent in Zarma must be used in
their french-ized forms (for better understandability).
\item \textbf{Prioritize understandability:} guidelines to prioritize understandability and fidelity over
word-for-word translation. We instructed annotators to focus on creating translations that sound
natural and widely understandable by Zarma speakers.
\item \textbf{Language specific constraints:} language specific guidelines that cannot be generalized.
\end{itemize}

The pipeline speeds up the process while maintaining the quality, since some of the outputs from the automatic stages were already correct.

\paragraph{Zarma Community Attitude Towards Tech}:
The Zarma community---and the whole Niger in general---are very open minded regarding
technology. When we started our very first resource creation for the Zarma language, we
received positive feedback and even help from the community, as long as we developed an
openly accessible solution for the community. \textbf{For the SMOL annotation, that trust helps us to
receive valuable help.} For instance, a government based institution verbally promised to
accompany any language preservation—machine learning focus in our case---if the outcome
will be open-sourced for community usage.

\subsection{Post-Edits for Moor\'e}
\paragraph{Annotation Pipeline}
The annotation process for Moor\'e did not involve any automated components; everything was
annotated by humans. The annotation focuses more on the guidelines provided by \smol{}, in
addition to some more as in the Zarma case.

\paragraph{Moor\'e Community Perspective}
The Moor\'e community, similarly to the zarma community, are very open minded towards technology;
especially if it touches cultural/language preservation. One main feedback we received from
some elders (parents of one of the annotators) was a warning to ONLY USE standard Moor\'e
orthography, not any equivalent. They want the language to be well documented according to
the language standards.

\subsection{Post-Edits for Indonesian}

A volunteer team post-edited translations of \texttt{smoldoc} and \texttt{smolsent} datasets that had done by Gemini 2.5 Pro.
The contributors to translation and post-editing were
Muhammad Ravi Shulthan Habibi,
David Anugraha, and
Genta Indra Winata. The post-edits resulted in about 70\% of the machine translations being changed.

Translators agreed that the system output was often too ``formal", ``stiff", or ``awkward". The ``formal" translations were furthermore not formal in an acceptible sense, but ``too awkward and stiff, even for a more formal situations", as an annotator said. Each word choice might be correct and standard in Indonesian, but when combined in a sentence, the result sounded unnatural. Therefore, the majority of the post-edits focused on making the translations sound more natural.

Nonetheless, overall the system output was already quite reasonable in terms of register. In some cases, though, it leaned toward being too rigid. The post-edits tried to loosen that into a consistent ``medium'' range, but with some flexibility depending on the style of each sentence (sometimes slightly more formal, sometimes slightly less) so the overall text still feels natural and coherent.

\subsection{Translations for Languages of the Russian Federation}
Traditionally, speakers of hundreds of Cyrillic-based languages in the Russian Federation translate datasets via Russian. For the success of this project, I (Ali Kuzhuget) first funded a professional translation into Russian, engaging Andrey Anisimov as the main translator. The proofreading was conducted by Farhad Fatkullin, Vice-President of the National League of Translators, together with machine translation specialist David Dal\'e. I also oversaw formatting correctness and coordinated the overall translation workflow.

In parallel, I supervise the translation of the dataset from Russian into Tuvan, using a dedicated Telegram chatbot for large-scale dataset translation. This tool enables multiple rounds of validation and systematic assessment of translation quality. Currently, representatives of about a dozen Cyrillic languages are in the process of translating the \smol{} dataset into their own languages through Russian and/or English (for example, Tuvan, Bashkir, Chuvash, and others), ensuring both linguistic accuracy and cultural relevance.

\section{Full results}
\label{app:fullresults}
\label{app:relatedlangs}

Full per-language results can be seen in Table \ref{tab:full-results}. Results are sorted by the $\Delta_{FT}$, which is the \chrf{} of the \smolboth{} model minus the \chrf{} of the finetuned \textsc{Both} model---in other words, how much the finetuning on \smol{} improved the baseline model.

\paragraph{Google Translate Languages and their cousins} As mentioned in the results section, some languages see only very small improvements from finetuning on \smol{}, and others even see losses. These are mainly either high-resource languages, or close relatives to higher-resource languages. In the full table \ref{tab:full-results} below, The superscript $^{{\tiny \textrm{GTr}}}$ indicates a language supported by Google Translate at the time of these experiments; a superscript like $^{\sim\textrm{xx}}$ means that this language is closely related to the Google-Translate-supported language xx. We only consider the 108 languages that were present on Google Translate at the time of this work.

\onecolumn
{ \scriptsize  
\begin{longtable}{l|l|r|rrrrr|rr|rrrr}

lang & cat & $\Delta_{FT}$ & G2F & +sS  & +sD & +sB  & +sG  & Cld & +RAG &  G2P & GPT4o & GTr & NLLB  \\
\hline
\endhead
ee & \textsc{Both} & +36.1 & 3.0 & 37.7 & 37.6 & 39.1 & 39.2 & 37.8 & 40.0 & 39.5 & 7.5 & { \bf 42.7 } & 40.7  \\
kr & \textsc{Both} & +10.8 & 17.3 & 25.6 & 25.9 & 28.1 & 28.8 & 22.7 & 26.3 & 20.3 & 22.2 & { \bf 32.6 } & 31.0  \\
kg & \textsc{Both} & +9.2 & 34.9 & 46.9 & 36.8 & 44.1 & 43.2 & 43.2 & 47.0 & 37.8 & 29.1 & { \bf 50.2 } & 3.4  \\
bem & \textsc{Both} & +7.3 & 40.0 & 44.8 & 44.7 & 47.3 & 49.2 & 43.3 & 47.7 & 42.3 & 33.3 & { \bf 49.7 } & 41.8  \\
dyu & \textsc{Both} & +5.3 & 17.9 & 22.5 & 23.3 & 23.2 & 23.7 & 23.9 & { \bf 24.4 } & 21.0 & 4.5 & 22.4 & 12.5  \\
din & \textsc{Both} & +4.6 & 20.3 & 23.8 & 22.9 & 24.9 & 25.1 & 23.3 & 25.9 & 21.4 & 1.6 & 25.1 & { \bf 26.5 }  \\
luo & \textsc{Both} & +4.1 & 37.4 & 39.1 & 41.1 & 41.5 & 42.0 & 39.1 & { \bf 42.0 } & 39.6 & 36.1 & 41.3 & 39.5  \\
fon & \textsc{Both} & +3.6 & 21.3 & 24.3 & 23.9 & 24.9 & 25.3 & 20.4 & 23.7 & 23.8 & 1.9 & { \bf 25.9 } & 24.2  \\
bm & \textsc{Both} & +3.4 & 30.8 & 28.6 & 35.2 & 34.2 & 34.1 & 34.0 & { \bf 36.2 } & 33.9 & 9.0 & 35.7 & 32.2  \\
ak & \textsc{Both} & +2.6 & 35.5 & 36.1 & 37.8 & 38.1 & { \bf 38.2 } & 34.4 & 38.1 & 37.3 & 32.2 & 34.5 & 33.3  \\
ln & \textsc{Both} & +2.5 & 46.8 & 48.1 & 48.4 & 49.3 & { \bf 49.3 } & 44.6 & 48.3 & 46.5 & 45.2 & 46.4 & 45.7  \\
wo & \textsc{Both} & +1.1 & 30.3 & 30.0 & 30.4 & 31.4 & 31.6 & 31.4 & 32.2 & 30.7 & 29.8 & { \bf 36.2 } & 30.9  \\
ff & \textsc{Both} & +0.9 & 25.0 & 24.4 & 26.4 & 25.9 & 26.5 & 25.7 & 26.1 & 25.2 & 2.5 & 25.9 & { \bf 27.1 }  \\
om & \textsc{Both} & -0.8 & 40.1 & 38.0 & 39.0 & 39.3 & 39.4 & 39.0 & 40.2 & 41.3 & 38.4 & { \bf 41.4 } & 39.1  \\
lg & \textsc{Both} & -1.1 & 42.5 & 39.9 & 41.4 & 41.4 & 41.7 & 42.0 & 43.1 & 43.5 & 41.0 & { \bf 43.6 } & 41.1  \\
ber & \textsc{Both} & -3.2 & 25.3 & 20.6 & 22.9 & 22.1 & 21.9 & 28.5 & 25.2 & 31.1 & 2.8 & 21.0 & { \bf 32.4 }  \\
\hdashline
trp & \textsc{SmolDoc} & +29.7 & 8.4 & 6.5 & 37.8 & 38.1 & { \bf 39.1 } & 24.7 & 35.8 & 27.2 & 20.3 & 35.9 & -  \\
mni-M. & \textsc{SmolSent} & +26.4 & 2.9 & 30.0 & 1.2 & 29.3 & 29.3 & 29.6 & 31.8 & 33.6 & 1.3 & { \bf 45.6 } & 0.8  \\
gaa & \textsc{Both} & +23.1 & 22.7 & 44.5 & 44.4 & 45.8 & 47.4 & 34.7 & 44.0 & 40.9 & 6.6 & { \bf 48.3 } & -  \\
dov & \textsc{Both} & +21.1 & 19.1 & 39.2 & 38.3 & 40.2 & 40.6 & 19.2 & 39.5 & 18.2 & 8.7 & { \bf 41.7 } & -  \\
ahr$^{\sim \textrm{hi}}$ & neither & +17.8 & 24.2 & 31.8 & 41.9 & 42.0 & { \bf 42.8 } & 32.8 & 39.0 & 30.0 & 36.9 & - & -  \\
sus & \textsc{Both} & +17.8 & 11.3 & 28.3 & 26.8 & 29.1 & 30.3 & 26.1 & 29.4 & 20.7 & 5.6 & { \bf 34.6 } & -  \\
nqo & \textsc{Both} & +17.5 & 0.2 & 17.9 & 17.1 & 17.7 & 17.5 & 17.1 & 17.9 & 17.2 & 1.1 & { \bf 19.1 } & -  \\
alz & \textsc{Both} & +15.5 & 16.9 & 31.5 & 30.5 & 32.4 & 33.4 & 25.3 & 30.6 & 26.9 & 8.9 & { \bf 36.6 } & -  \\
lu & \textsc{Both} & +13.8 & 27.6 & 37.5 & 39.3 & 41.4 & { \bf 42.2 } & 27.2 & 37.0 & 34.8 & 21.9 & - & -  \\
cgg & \textsc{Both} & +12.2 & 32.6 & 40.5 & 40.5 & 44.8 & { \bf 44.8 } & 37.3 & 42.2 & 37.7 & 28.3 & 42.8 & -  \\
ks-D.$^{\sim \textrm{hi}}$ & neither & +11.7 & 14.9 & 26.6 & 18.8 & 26.6 & { \bf 27.7 } & 23.8 & 27.1 & 21.5 & 19.2 & - & 21.1  \\
brx & \textsc{SmolSent} & +11.6 & 24.3 & 36.0 & 0.4 & 35.9 & { \bf 37.0 } & 30.8 & 35.9 & 36.2 & 5.2 & - & -  \\
mag$^{\sim \textrm{hi}}$ & neither & +8.1 & 47.0 & 45.3 & 55.7 & 55.1 & 54.7 & 47.3 & 51.8 & 47.4 & 48.7 & - & { \bf 59.4 }  \\
ki & \textsc{Both} & +7.7 & 32.6 & 38.2 & 38.0 & 40.3 & 40.6 & 35.9 & { \bf 42.0 } & 39.1 & 10.5 & - & 38.4  \\
aa & \textsc{Both} & +7.4 & 14.2 & 20.1 & 20.3 & 21.6 & 21.8 & 18.9 & 20.6 & 18.9 & 5.6 & { \bf 23.1 } & -  \\
ks$^{\sim \textrm{ur}}$ & neither & +7.3 & 22.1 & 29.4 & 0.4 & 29.4 & 29.7 & 28.0 & 30.4 & 30.5 & 26.3 & - & { \bf 36.7 }  \\
nr$^{\sim \textrm{zu}}$ & neither & +7.0 & 48.9 & 54.0 & 51.1 & 55.9 & 57.5 & 48.0 & 53.6 & 51.2 & 45.5 & { \bf 59.5 } & -  \\
doi$^{\sim \textrm{hi}}$ & neither & +6.6 & 34.3 & 28.1 & { \bf 41.4 } & 40.9 & 41.3 & 35.9 & 39.5 & 38.2 & 27.7 & 40.4 & -  \\
sat-L. & \textsc{SmolSent} & +6.4 & 12.8 & 19.5 & 15.5 & 19.2 & 20.8 & { \bf 25.3 } & 22.5 & 22.7 & 21.3 & 22.7 & -  \\
mfe$^{\sim \textrm{fr}}$ & neither & +5.3 & 59.5 & 65.4 & 62.6 & 64.8 & 66.9 & 59.6 & 65.0 & 59.8 & 59.5 & { \bf 67.5 } & -  \\
ach & \textsc{Both} & +5.2 & 33.2 & 43.1 & 32.5 & 38.4 & 39.2 & 32.4 & 37.3 & 35.1 & 23.8 & { \bf 43.2 } & -  \\
ayl$^{\sim \textrm{ar}}$ & neither & +4.5 & 47.3 & 51.7 & 51.9 & 51.8 & { \bf 53.9 } & 45.3 & 48.9 & 46.3 & 48.6 & - & -  \\
st$^{{\tiny \textrm{GTr}}}$ & neither & +4.5 & 49.9 & 54.0 & 55.2 & 54.4 & 55.0 & 49.4 & { \bf 57.0 } & 53.1 & 49.2 & 49.0 & 47.2  \\
ber-L. & \textsc{Both} & +4.2 & 26.1 & 27.9 & 30.3 & 30.3 & 30.9 & 27.6 & 32.7 & 32.1 & 21.2 & { \bf 34.7 } & -  \\
apd-S.$^{\sim \textrm{ar}}$ & neither & +3.6 & 42.3 & { \bf 50.2 } & 43.3 & 45.9 & 47.1 & 43.2 & 45.0 & 42.5 & 45.6 & - & -  \\
ve$^{\sim \textrm{sn}}$ & neither & +3.5 & 47.9 & 50.0 & 48.7 & 51.4 & 52.2 & 50.2 & 53.1 & 52.7 & 43.9 & { \bf 56.8 } & -  \\
kri$^{\sim \textrm{en}}$ & neither & +2.8 & 31.5 & 34.2 & 31.8 & 34.3 & 34.7 & 34.5 & 33.5 & 30.7 & 34.9 & { \bf 34.9 } & -  \\
tiv & \textsc{Both} & +2.5 & 23.8 & 25.7 & 25.8 & 26.3 & { \bf 26.5 } & 22.3 & 24.2 & 24.5 & 1.5 & 25.2 & -  \\
gn & \textsc{SmolSent} & +2.3 & 37.4 & 37.8 & 30.4 & { \bf 39.7 } & 38.4 & 36.0 & 38.0 & 36.4 & 35.6 & 38.4 & 38.5  \\
mos & \textsc{Both} & +2.3 & 18.2 & 20.9 & 18.9 & 20.5 & 21.1 & 24.3 & { \bf 25.0 } & 20.9 & 1.3 & - & 23.8  \\
tum$^{\sim \textrm{ny}}$ & neither & +1.1 & 40.8 & 39.5 & 42.4 & 41.9 & 42.8 & 40.0 & 42.7 & 43.8 & 37.7 & { \bf 45.4 } & 36.2  \\
ti$^{\sim \textrm{am}}$ & neither & +0.8 & 24.2 & 24.3 & 24.9 & 25.0 & 25.7 & 25.0 & { \bf 26.2 } & 26.1 & 9.3 & 26.1 & 25.5  \\
yo$^{{\tiny \textrm{GTr}}}$ & neither & +0.6 & 34.6 & 33.8 & 35.4 & 35.2 & 35.8 & 29.2 & { \bf 36.8 } & 26.7 & 27.6 & 21.3 & 32.6  \\
tn$^{\sim \textrm{st}}$ & neither & +0.2 & 52.5 & 50.8 & 51.9 & 52.7 & 53.2 & 50.1 & 51.7 & 53.3 & 36.8 & { \bf 55.6 } & 53.0  \\
ar-M.$^{\sim \textrm{ar}}$ & neither & +0.1 & 40.1 & 41.8 & 39.1 & 40.2 & 40.9 & 40.5 & 40.8 & 40.4 & 41.0 & - & { \bf 43.0 }  \\
am$^{{\tiny \textrm{GTr}}}$ & neither & +0.1 & 34.0 & 33.2 & 33.0 & 34.1 & 33.5 & 31.6 & 32.6 & { \bf 35.8 } & 29.6 & 34.7 & 30.3  \\
ig$^{{\tiny \textrm{GTr}}}$ & neither & -0.1 & 47.2 & 47.1 & 47.0 & 47.1 & 47.8 & 43.9 & 46.2 & { \bf 47.8 } & 46.2 & 47.6 & 46.6  \\
so$^{{\tiny \textrm{GTr}}}$ & neither & -0.1 & 49.7 & 46.2 & 50.0 & 49.6 & 49.1 & 48.7 & 49.8 & 50.3 & { \bf 50.8 } & 50.6 & 48.6  \\
arz$^{\sim \textrm{ar}}$ & neither & -0.1 & 48.6 & 46.1 & 48.9 & 48.5 & 47.8 & 49.6 & { \bf 50.3 } & 48.8 & 49.7 & - & 49.6  \\
kl & \textsc{SmolSent} & -0.3 & 40.6 & 39.7 & 30.2 & 40.3 & 41.2 & 42.2 & { \bf 43.1 } & 41.2 & 41.6 & 42.9 & -  \\
sa & \textsc{SmolSent} & -0.3 & 33.0 & 32.9 & 26.7 & 32.7 & 33.4 & 31.8 & 33.0 & 32.1 & 32.0 & { \bf 35.2 } & 29.0  \\
ay & \textsc{SmolSent} & -0.5 & 32.7 & 31.8 & 24.2 & 32.2 & 32.4 & 33.4 & 33.2 & 32.9 & 30.0 & { \bf 34.7 } & 31.7  \\
sn$^{{\tiny \textrm{GTr}}}$ & neither & -0.5 & 50.5 & 48.3 & 50.2 & 50.0 & 50.3 & 46.8 & 48.8 & { \bf 51.8 } & 50.3 & 49.2 & 48.2  \\
efi & \textsc{Both} & -0.6 & 14.7 & 14.5 & 14.3 & 14.1 & 14.2 & { \bf 15.3 } & 15.1 & 15.0 & 2.2 & - & -  \\
ss$^{\sim \textrm{zu}}$ & neither & -0.6 & 50.2 & 48.8 & 48.2 & 49.6 & 50.3 & 49.6 & 51.2 & 51.8 & 46.0 & { \bf 56.3 } & 48.1  \\
yue$^{\sim \textrm{zh}}$ & neither & -0.7 & 26.8 & 25.8 & 25.1 & 26.1 & 26.1 & 28.2 & 28.2 & 27.5 & { \bf 31.6 } & 25.9 & 22.6  \\
bci & \textsc{Both} & -0.7 & 23.2 & 22.2 & 21.8 & 22.5 & 22.9 & 17.1 & 20.7 & 27.6 & 1.0 & { \bf 29.8 } & -  \\
ndc-Z.$^{\sim \textrm{sn}}$ & neither & -1.0 & 29.2 & 27.9 & 28.9 & 28.2 & 28.3 & 27.6 & 28.0 & { \bf 29.6 } & 28.6 & 29.5 & -  \\
es$^{{\tiny \textrm{GTr}}}$ & neither & -1.1 & 62.4 & 61.3 & 51.6 & 61.3 & 61.3 & - & - & 63.0 & - & { \bf 63.5 } & 61.8  \\
sat & \textsc{SmolSent} & -1.3 & 32.4 & 30.9 & 1.0 & 31.1 & 30.8 & 34.7 & 36.0 & { \bf 36.3 } & 1.8 & 35.7 & -  \\
rw$^{{\tiny \textrm{GTr}}}$ & neither & -1.4 & 45.2 & 43.1 & 43.0 & 43.8 & 43.8 & 43.2 & 44.0 & 45.1 & 44.7 & { \bf 48.8 } & 43.4  \\
nd$^{\sim \textrm{zu}}$ & neither & -1.4 & 43.9 & 41.5 & 42.6 & 42.5 & 43.2 & 42.3 & 42.9 & { \bf 44.5 } & 43.6 & - & -  \\
sw$^{{\tiny \textrm{GTr}}}$ & neither & -1.5 & 66.7 & 64.6 & 64.2 & 65.2 & 64.8 & 64.0 & 65.5 & { \bf 67.2 } & 66.5 & 65.3 & 60.5  \\
mg$^{{\tiny \textrm{GTr}}}$ & neither & -1.9 & 52.8 & 48.9 & 51.6 & 50.9 & 51.5 & 52.4 & 52.5 & { \bf 53.3 } & 52.2 & 52.6 & 52.1  \\
qu & \textsc{SmolSent} & -1.9 & 34.7 & 32.8 & 30.4 & 32.8 & 33.0 & 35.3 & 35.1 & 34.0 & 22.0 & { \bf 36.3 } & 27.9  \\
zu$^{{\tiny \textrm{GTr}}}$ & neither & -2.0 & 58.3 & 56.4 & 55.3 & 56.3 & 56.1 & 54.1 & 55.5 & { \bf 58.5 } & 57.5 & 57.6 & 57.6  \\
lus & \textsc{SmolSent} & -2.1 & 42.6 & 39.7 & 38.0 & 40.5 & 41.4 & 40.6 & 41.5 & { \bf 43.8 } & 33.5 & 42.6 & 39.0  \\
scn$^{\sim \textrm{it}}$ & neither & -2.2 & 52.4 & 47.3 & 50.0 & 50.2 & 50.8 & 49.9 & 51.4 & 52.1 & 49.5 & { \bf 53.3 } & 51.0  \\
nso$^{\sim \textrm{st}}$ & neither & -2.3 & 46.8 & 42.8 & 43.6 & 44.5 & 44.6 & 46.9 & 47.7 & { \bf 48.1 } & 46.9 & 47.6 & 45.5  \\
xh$^{{\tiny \textrm{GTr}}}$ & neither & -2.3 & 53.9 & 49.7 & 50.7 & 51.6 & 51.8 & 51.6 & 52.3 & 53.9 & 53.7 & { \bf 54.8 } & 51.2  \\
ne$^{{\tiny \textrm{GTr}}}$ & neither & -2.7 & 54.3 & 51.8 & 51.7 & 51.6 & 52.1 & 52.4 & 52.5 & 52.4 & 52.7 & { \bf 54.9 } & 45.2  \\
pa-A.$^{\sim \textrm{pa}}$ & neither & -3.0 & 38.1 & 35.8 & 0.3 & 35.1 & 35.7 & 41.6 & 41.2 & 36.7 & 37.3 & { \bf 43.5 } & -  \\
aeb$^{\sim \textrm{ar}}$ & neither & -3.3 & 46.5 & 41.9 & 42.3 & 43.2 & 43.4 & 45.9 & 46.8 & 47.6 & { \bf 49.2 } & - & 43.8  \\
ha$^{{\tiny \textrm{GTr}}}$ & neither & -3.4 & { \bf 54.5 } & 50.7 & 49.9 & 51.1 & 51.5 & 50.9 & 51.0 & 54.1 & 53.9 & 53.8 & 53.9  \\
ts$^{\sim \textrm{zu}}$ & neither & -3.6 & 50.6 & 47.2 & 46.4 & 47.0 & 48.1 & 49.7 & 50.1 & 51.6 & 49.0 & { \bf 52.9 } & 51.3  \\
rn$^{\sim \textrm{rw}}$ & neither & -3.9 & 44.5 & 39.3 & 40.5 & 40.6 & 40.9 & 43.1 & 43.4 & { \bf 46.2 } & 44.3 & 45.4 & 45.0  \\
af$^{{\tiny \textrm{GTr}}}$ & neither & -4.2 & 71.9 & 68.8 & 67.9 & 67.7 & 68.3 & 71.7 & { \bf 72.5 } & 72.1 & 71.8 & 71.5 & 68.6  \\
bo & \textsc{SmolSent} & -4.3 & 41.3 & 36.7 & 34.7 & 37.0 & 37.3 & 42.6 & 42.1 & { \bf 43.3 } & 19.8 & 41.8 & 36.9  \\
ny$^{{\tiny \textrm{GTr}}}$ & neither & -5.1 & 55.0 & 47.5 & 50.5 & 49.9 & 49.7 & 53.0 & 53.1 & 55.3 & 53.9 & { \bf 55.8 } & 50.3  \\
pcm$^{\sim \textrm{en}}$ & neither & -6.5 & 47.9 & 43.5 & 39.4 & 41.4 & 41.6 & 51.3 & 45.7 & 49.8 & { \bf 56.0 } & - & -  \\
tcy & \textsc{SmolDoc} & -6.8 & 34.7 & 22.0 & 28.1 & 27.9 & 28.8 & 28.2 & 29.3 & 36.7 & 21.6 & { \bf 39.1 } & -  \\
ktu & \textsc{Both} & -9.4 & 56.6 & 59.3 & 40.4 & 47.2 & 51.3 & 45.8 & 48.4 & 57.8 & 22.3 & { \bf 64.3 } & - \\
\hline
\caption{Full results (0-shot) For the \enxx{} direction. Languages in the Intersect subset (supported by all models) are shown first, and then all other languages. The $\Delta_{FT}$ compares the base model and the model finetuned on \smolboth{}, to give an idea of how effective the \smol{} datasets are for that language. The \textsc{cat} column indicates which \smol{} datasets support this language. The superscript ${\tiny \textrm{GTr}}$ indicates a language supported by Google Translate; a superscript like $\sim\textrm{xx}$ means that this language is closely related to that Google-Translate-supported language. \\\\ \textbf{Abbreviations:} This table needed some squishing to fit. Language varieties whose script/region is different from the CLDR default would have the ISO-15924 script code in the BCP-47 code, like \textsc{mni-Mtei} or \textsc{ber-Latn}; in this table we have abbreviated them to the first letter thereof (\textsc{mni-M} or \textsc{ber-L}). Similarly, we have abbreviated: \\ \smolseal{} $\rightarrow$ \textsc{sS} \\
\smoldog{} $\rightarrow$ \textsc{sD} \\
\smolboth{} $\rightarrow$ \textsc{sB} \\
\textsc{Both+Gatitos} $\rightarrow$ \textsc{sG} \\
\textsc{Gemini 2.0-\{Flash, Pro\}} $\rightarrow$ \textsc{G2.0-\{F,P\}} \\
\textsc{Google Translate} $\rightarrow$ \textsc{GTr}. }
\label{tab:full-results}
\end{longtable}
}

\twocolumn

\section{Complete Per-Language details: the Big-\smol{} table}
A summary of all \smol{} language pairs and coarse-grained information about them can be seen in Table \ref{tab:full-details}. Numbers are given in terms of examples; keep in mind that a single example in \smoldog{} is a document, whereas in \smolseal{} it is a sentence.

\onecolumn
{ \scriptsize
\begin{longtable}{llllrrr}

Lang. pair	& target language name   & ISO 15924 Script &Continent & trg.chars &	\textsc{S.Doc} &	\textsc{S.Sent} \\
\hline
\endhead
en\_yo & Yoruba & Latn & Africa & 780k & 584 & 863 \\
en\_sw & Swahili & Latn & Africa & 699k & 584 & 863 \\
en\_ha & Hausa & Latn & Africa & 696k & 584 & 863 \\
en\_grt-Latn & Garo (Latin script) & Latn & Asia & 591k & 457 & 0 \\
en\_trp & Kokborok & Latn & Asia & 581k & 457 & 0 \\
en\_mg & Malagasy & Latn & Africa & 580k & 391 & 863 \\
en\_xsr-Tibt & Sherpa (Tibetan script) & Tibt & Asia & 569k & 457 & 0 \\
en\_om & Oromo & Latn & Africa & 542k & 391 & 863 \\
en\_sd-Deva & Sindhi (Devanagari script) & Deva & Asia & 525k & 456 & 0 \\
en\_ccp-Latn & Chakma (Latin script) & Latn & Asia & 521k & 457 & 0 \\
en\_spv & Sambalpuri & Orya & Asia & 508k & 457 & 0 \\
en\_doi & Dogri & Deva & Asia & 503k & 454 & 0 \\
en\_xnr & Kangri & Deva & Asia & 503k & 457 & 0 \\
en\_mjl & Mandeali & Deva & Asia & 496k & 457 & 0 \\
en\_lif-Limb & Limbu (Limbu script) & Limb & Asia & 494k & 457 & 0 \\
en\_ne & Nepali & Deva & Asia & 494k & 456 & 0 \\
en\_kru & Kurukh & Deva & Asia & 492k & 457 & 0 \\
en\_hoc-Wara & Ho (Warang Chiti script) & Wara & Asia & 492k & 457 & 0 \\
en\_bra & Braj & Deva & Asia & 491k & 457 & 0 \\
en\_bns & Bundeli & Deva & Asia & 490k & 456 & 0 \\
en\_mag & Magahi & Deva & Asia & 488k & 456 & 0 \\
en\_wbr & Wagdi & Deva & Asia & 488k & 455 & 0 \\
en\_bfy & Bagheli & Deva & Asia & 487k & 457 & 0 \\
en\_unr-Deva & Mundari (Devanagari script) & Deva & Asia & 485k & 457 & 0 \\
en\_mtr & Mewari & Deva & Asia & 480k & 457 & 0 \\
en\_tcy & Tulu & Knda & Asia & 480k & 451 & 0 \\
en\_ahr & Ahirani & Deva & Asia & 479k & 457 & 0 \\
en\_ig & Igbo & Latn & Africa & 474k & 391 & 863 \\
en\_dhd & Dhundari & Deva & Asia & 465k & 456 & 0 \\
en\_bfq & Badaga & Taml & Asia & 464k & 457 & 0 \\
en\_kfy & Kumaoni & Deva & Asia & 462k & 457 & 0 \\
en\_bgq & Bagri & Deva & Asia & 462k & 457 & 0 \\
en\_scl & Shina & Arab & Asia & 460k & 457 & 0 \\
en\_am & Amharic & Ethi & Africa & 443k & 584 & 863 \\
en\_lep & Lepcha & Lepc & Asia & 441k & 456 & 0 \\
en\_st & Sesotho & Latn & Africa & 412k & 260 & 863 \\
en\_sgj & Surgujia & Deva & Asia & 395k & 356 & 0 \\
en\_so & Somali & Latn & Africa & 392k & 260 & 862 \\
en\_ny & Chichewa & Latn & Africa & 386k & 260 & 863 \\
en\_sn & Shona & Latn & Africa & 382k & 260 & 863 \\
en\_rw & Kinyarwanda & Latn & Africa & 378k & 260 & 863 \\
en\_zu & Zulu & Latn & Africa & 373k & 260 & 863 \\
en\_lg & Luganda & Latn & Africa & 369k & 260 & 863 \\
en\_xh & Xhosa & Latn & Africa & 368k & 260 & 863 \\
en\_ln & Lingala & Latn & Africa & 365k & 260 & 863 \\
en\_noe & Nimadi & Deva & Asia & 342k & 315 & 0 \\
en\_luo & Luo & Latn & Africa & 340k & 260 & 863 \\
en\_bm & Bambara & Latn & Africa & 337k & 260 & 863 \\
en\_ak & Twi & Latn & Africa & 328k & 260 & 863 \\
en\_sjp & Surjapuri & Deva & Asia & 327k & 299 & 0 \\
en\_wo & Wolof & Latn & Africa & 321k & 260 & 863 \\
en\_ff & Fulani & Latn & Africa & 320k & 260 & 862 \\
sw\_ar & Arabic & Arab & Asia & 274k & 330 & 0 \\
en\_ar-MA & Morrocan Arabic & Arab & Africa & 273k & 260 & 863 \\
en\_arz & Egyptian Arabic & Arab & Africa & 265k & 260 & 863 \\
am\_ar & Arabic & Arab & Asia & 265k & 329 & 0 \\
en\_nso & Sepedi & Latn & Africa & 243k & 130 & 863 \\
en\_ti & Tigrinya & Ethi & Africa & 231k & 260 & 863 \\
en\_af & Afrikaans & Latn & Africa & 219k & 130 & 863 \\
en\_ber-Latn & Tamazight (Latin Script) & Latn & Africa & 206k & 130 & 862 \\
en\_ber & Tamazight (Tifinagh Script) & Tfng & Africa & 206k & 130 & 862 \\
en\_ee & Ewe & Latn & Africa & 202k & 130 & 863 \\
en\_pcm & Nigerian Pidgin & Latn & Africa & 195k & 130 & 863 \\
en\_yue & Cantonese & Hant & Asia & 195k & 584 & 863 \\
en\_kri & Krio & Latn & Africa & 188k & 130 & 863 \\
en\_tn & Tswana & Latn & Africa & 182k & 66 & 863 \\
en\_ve & Venda & Latn & Africa & 167k & 66 & 863 \\
en\_bm-Nkoo & NKo & Nkoo & Africa & 167k & 66 & 863 \\
en\_bem & Bemba (Zambia) & Latn & Africa & 166k & 66 & 863 \\
en\_ts & Tsonga & Latn & Africa & 165k & 66 & 863 \\
en\_tum & Tumbuka & Latn & Africa & 164k & 66 & 863 \\
en\_ss & Swati & Latn & Africa & 163k & 66 & 863 \\
en\_ktu & Kituba (DRC) & Latn & Africa & 162k & 66 & 863 \\
en\_nr & South Ndebele & Latn & Africa & 159k & 66 & 863 \\
en\_lij & Ligurian & Latn & Europe & 158k & 66 & 863 \\
en\_fon & Fon & Latn & Africa & 157k & 66 & 863 \\
en\_id & Indonesian & Latn & Asia & 157k & 66 & 863 \\
en\_ndc-ZW & Ndau & Latn & Africa & 156k & 66 & 863 \\
en\_kg & Kongo & Latn & Africa & 154k & 66 & 863 \\
en\_dov & Dombe & Latn & Africa & 153k & 66 & 863 \\
en\_nd & North Ndebele & Latn & Africa & 150k & 66 & 863 \\
en\_ki & Kikuyu & Latn & Africa & 149k & 66 & 863 \\
en\_lu & Kiluba (Luba-Katanga) & Latn & Africa & 148k & 66 & 863 \\
en\_efi & Efik & Latn & Africa & 147k & 66 & 863 \\
en\_cgg & Kiga & Latn & Africa & 147k & 66 & 863 \\
en\_din & Dinka & Latn & Africa & 145k & 66 & 863 \\
en\_rn & Rundi & Latn & Africa & 144k & 66 & 863 \\
en\_tiv & Tiv & Latn & Africa & 141k & 66 & 863 \\
en\_kr & Kanuri & Latn & Africa & 139k & 66 & 863 \\
en\_alz & Alur & Latn & Africa & 139k & 66 & 863 \\
en\_mfe & Mauritian Creole & Latn & Africa & 137k & 66 & 863 \\
en\_dyu & Dyula & Latn & Africa & 136k & 66 & 863 \\
en\_ach & Acholi & Latn & Africa & 135k & 66 & 863 \\
en\_dje & Zarma & Latn & Africa & 135k & 66 & 863 \\
en\_aa & Afar & Latn & Africa & 133k & 66 & 863 \\
en\_bci & Baoulé & Latn & Africa & 131k & 66 & 863 \\
en\_sus & Susu & Latn & Africa & 128k & 66 & 863 \\
en\_gaa & Ga & Latn & Africa & 126k & 66 & 863 \\
en\_mos & Mooré & Latn & Africa & 125k & 66 & 863 \\
en\_aeb & Tunisian Arabic & Arab & Africa & 115k & 66 & 862 \\
en\_apd & Sudanese Arabic & Arab & Africa & 112k & 66 & 855 \\
en\_ayl & Libyan Arabic & Arab & Africa & 109k & 66 & 863 \\
en\_scn & Sicilian & Latn & Europe & 102k & 100 & 0 \\
sw\_zh & Mandarin Chinese & Hans & Asia & 101k & 330 & 0 \\
en\_kl & Kalaallisut & Latn & Americas & 97k & 0 & 863 \\
am\_zh & Mandarin Chinese & Hans & Asia & 96k & 329 & 0 \\
en\_es & Spanish & Latn & Europe & 88k & 0 & 863 \\
en\_sat & Santali (Ol Chiki script) & Olck & Asia & 83k & 0 & 863 \\
en\_bo & Tibetan & Tibt & Asia & 82k & 0 & 863 \\
en\_lus & Mizo & Latn & Asia & 82k & 0 & 863 \\
en\_gn & Guarani & Latn & Americas & 82k & 0 & 863 \\
en\_ay & Aymara & Latn & Americas & 82k & 0 & 863 \\
en\_sat-Latn & Santali (Latin Script) & Latn & Asia & 81k & 0 & 863 \\
en\_is & Icelandic & Latn & Europe & 77k & 0 & 863 \\
en\_hac & Hawrami & Arab & Asia & 77k & 0 & 863 \\
en\_sa & Sanskrit & Deva & Asia & 77k & 0 & 863 \\
en\_qu & Quechua & Latn & Americas & 74k & 0 & 863 \\
en\_brx & Bodo (India) & Deva & Asia & 74k & 0 & 863 \\
en\_ckb & Kurdish (Sorani) & Arab & Asia & 73k & 0 & 863 \\
en\_ks & Kashmiri & Arab & Asia & 73k & 0 & 863 \\
en\_pa-Arab & Lahnda Punjabi (Pakistan) & Arab & Asia & 73k & 0 & 863 \\
en\_mni-Mtei & Meiteilon (Manipuri) & Mtei & Asia & 71k & 0 & 863 \\
en\_ks-Deva & Kashmiri (Devanagari script) & Deva & Asia & 65k & 0 & 863 \\
en\_mhr & Meadow Mari & Cyrl & Asia & 63k & 66 & 0 \\
en\_glk & Gilaki & Arab & Asia & 52k & 0 & 863 \\
\hline
\caption{Details on all \smol{} language pairs, sorted by the total number of characters in the target side (col. 5). The last two columns are the number of examples per language pair; keep in mind that an example for \smolseal{} is a sentence pair but for \smoldog{} is a document/paragraph. Language pairs are only listed in the direction in which they were translated, so no \xxen{} pairs are present.}
\label{tab:full-details}
\end{longtable}
}


\section{Data sample}

\subsection{Sample datum from SmolSent}
\begin{lstlisting}

{'id': 381,
 'sl': 'en',
 'tl': 'luo',
 'is_src_orig': True,
 'src': 'Rih, a deaf former soldier, plots rebellion while married to a queer, teenage god.',
 'trg': 'Rih, mane en jalweny ma Radin, ochano balo ka koni to okendo ng'ano manigi kido mar chuech kamare, nyasaye ma en ojana.'
}
\end{lstlisting}

\subsection{Sample datum from SmolDoc}
\begin{lstlisting}
{
'id': 'topic_587__weyiwiniwaaotiwenwy',
 'sl': 'en',
 'tl': 'pcm',
 'is_src_orig': True,
 'factuality': 'ok',  # this is a story so there is no factual claim that could be wrong
 'srcs': ['"What the hell are you doing, you idiot?!"',
          '"Excuse me?"',
          '"You cut me off! You almost made me crash!"',
          '"I'm sorry, I didn't mean to. I was just trying to get around that slow-moving truck."',
          '"Well, you could have at least used your turn signal!"',
          '"I did use my turn signal!"',
          '"No, you didn't! You just pulled right out in front of me!"',
          '"I'm telling you, I used my turn signal!"',
          '"Whatever. You're still a terrible driver."',
          '"And you're a jerk!"',
          '"At least I know how to drive!"',
          '"Oh, yeah? Well, I'm a better writer than you are!"',
          '"That's debatable."',
          '"It's not debatable! I'm Ernest Hemingway!"',
          '"Who?"',
          '"Ernest Hemingway! The greatest writer of all time!"',
          '"Never heard of him."',
          '"Well, you've heard of me now!"',
          '"Yeah, I heard of you."'],
 'trgs': ['"Wetin di hell dey do, yu idiot?!"',
          '"Ekskuse mi?"',
          '"Yu komot mi! Yu almost make mi krash!"',
          '"I dey sorry, I nor wont do am. I just dey try get around dat truk wey slow."',
          '"Well, yu for don yus yor turn sign!"',
          '"I yus mai turn sign!"',
          '"No, yu nor turn am! Yu just turn rite in front of mi!"',
          '"I dey tell yu, I yus mai turn sign!"',
          '"Wateva. Yu still bi one tribol driva."',
          '"And yu bi jerk!"',
          '"At least I sabi hau to drive!"',
          '"Oh, yeah? Well, I bi ogbonge writa pass yu!"',
          '"Wi fit dibate dat."',
          '"nortin to dibate! I bi Ernest Hemingway!"',
          '"Who?"',
          '"Ernest Hemingway! De writa of all taim wey grate pass!"',
          '"Neva hear am."',
          '"Well, yu don hear mi nau!"',
          '"Na so, I don hear yu."']
}
\end{lstlisting}

\end{document}